\RequirePackage{fix-cm}
\documentclass[smallextended]{svjour3}       
\smartqed  
\usepackage{graphicx}
\usepackage[center]{caption}
\usepackage{graphicx} 
\usepackage{multirow}
\usepackage{amsmath}
\usepackage{amsfonts}
\usepackage{amssymb}
\usepackage{algorithm}
\usepackage{algorithmic}
\usepackage{subfigure}
\usepackage[table]{xcolor} 
\begin{document}

\title{Uncertainty Quantification in Large Language Models Through Convex Hull Analysis}
\titlerunning{Uncertainty Quantification in Large Language Models}


\author{Ferhat Ozgur Catak \and Murat Kuzlu}


\institute{F.O. Catak\at
              Department of Electrical Engineering and Computer Science, University of Stavanger, Rogaland, Norway\\
              \email{f.ozgur.catak@uis.no}           
              \\
M. Kuzlu \at
              Department of Engineering Technology, Old Dominion University, Norfolk, VA, USA \\
              \email{mkuzlu@odu.edu} 
}

\date{Received: date / Accepted: date}

\maketitle

\begin{abstract}
Uncertainty quantification approaches have been more critical in large language models (LLMs), particularly high-risk applications requiring reliable outputs. However, traditional methods for uncertainty quantification, such as probabilistic models and ensemble techniques, face challenges when applied to the complex and high-dimensional nature of LLM-generated outputs. This study proposes a novel geometric approach to uncertainty quantification using convex hull analysis. The proposed method leverages the spatial properties of response embeddings to measure the dispersion and variability of model outputs. The prompts are categorized into three types, i.e., `easy', `moderate', and `confusing', to generate multiple responses using different LLMs at varying temperature settings. The responses are transformed into high-dimensional embeddings via a BERT model and subsequently projected into a two-dimensional space using Principal Component Analysis (PCA). The Density-Based Spatial Clustering of Applications with Noise (DBSCAN) algorithm is utilized to cluster the embeddings and compute the convex hull for each selected cluster. The experimental results indicate that the uncertainty of the model for LLMs depends on the prompt complexity, the model, and the temperature setting. 
\end{abstract}

\section{Introduction}
In recent years, with advanced computing technologies, the field of Natural Language Processing (NLP) has improved significantly, leading to the development of sophisticated large language models (LLM), such as GPT-3.5/4, Gemini, LLama, and others \cite{bharathi2024analysis,bsharat2023principled}. These LLMs have increasingly been used as powerful tools for NLP applications, and demonstrated significant performance in generating contextually relevant text and supporting a variety of applications, such as chatbots, automated content generation, virtual agents, text categorization, language translation, and many more \cite{suta2020overview,kandpal2020contextual,javaid2023chatgpt}.


%
In the literature, the number of studies in NLP and LLMs has dramatically increased after introducing the Generative Pre-trained Transformer 3 (GPT-3) by OpenAI in 2020, which was the most sophisticated neural network by that time \cite{aydin2022research}. It could create enhanced textual, visual, and auditory content without human intervention \cite{10155107}. However, there are serious concerns regarding LLMs' responses in terms of reliability and uncertainty. The authors in \cite{abdar2021review} highlight the importance of Uncertainty Quantification (UQ) methods in reducing uncertainties in optimization and decision-making processes with traditional methods, such as Bayesian approximation and ensemble learning. The authors in \cite{felicioni2024importance} emphasize the role of uncertainty estimation, particularly epistemic uncertainty, for LLMs, and conduct different techniques, i.e., Laplace Approximation, Dropout, and Epinets. They indicate that uncertainty plays a crucial role in developing LLM-based agents for decision making, and its information can significantly contribute to the overall performance of LLMs. However, these traditional approaches have difficulties dealing with the complex and high-dimensional characteristics of texts generated by LLMs. Therefore, novel approaches are needed to effectively capture the uncertainty in  LLMs' responses and enhance their trustworthiness. \cite{zhang2024luq}. The authors in \cite{huang2023look} conduct an early exploratory study of uncertainty measurement with twelve uncertainty estimation methods to help characterize the prediction risks of LLMs, as well as to investigate the correlations between LLMs’ prediction uncertainty and their performance. Their results indicate that there is a high potential for uncertainty estimation to identify the inaccurate or unreliable predictions generated by LLMs. To quantify the uncertainty of LLM,  the authors propose two metrics, i.e., Verbalized Uncertainty and Probing Uncertainty in \cite{tanneru2024quantifying}. Verbalized uncertainty employs prompting the LLM to express its confidence in its explanations, whereas probing uncertainty employs sample and model perturbations to quantify the uncertainty. According to the results, the probing uncertainty offers a better performance than verbalized uncertainty with lower uncertainty corresponding to explanations with higher faithfulness. The study \cite{huang2024uncertainty} presents a framework, called Rank-Calibration, to assess uncertainty and confidence measures for LLMs. The framework also provides a comprehensive robustness analysis and interpretability for LLMs.  The other study \cite{yang2023improving} also presents a framework to improve the reliability of LLMs, called Uncertainty-Aware In-context Learning. It involves fine-tuning the LLM using a calibration dataset, and evaluates the model’s knowledge by analyzing multiple responses to the same query to determine if a correct answer is present. 



To address the gap, in this paper, a convex hull-based geometric approach is proposed for UQ of LLMs. The main contributions of this study are (1) Introduce a geometric approach to UQ in LLMs' responses, (2) Implement a comprehensive system processing a wide range of prompts, and generating responses using different models and temperature settings, and (3) Highlight the relationship between the prompt complexity, model settings, and uncertainty.
    


\section{Related Work}
In recent years, UQ has been one of the challenging topics in the context of AI/ML concept, particularly critical applications using LLMs in healthcare, finance and law \cite{chen2024survey,ouyang2024ethical,savage2024large,catak2024trustworthy,cheong2024not}. Uncertainty can originate from different sources, including model architecture, model parameters, and dataset(s) \cite{nemani2023uncertainty,stracuzzi2018quantifying,jalaian2019uncertain,pernot2023calibration} in addition to the nature of LLMs. Training data can also be an influential source of uncertainty due to complexity and diversity of the selected dataset(s). The study \cite{abdar2021review}  provides a comprehensive overview of the UQ methods. Regarding UQ, there are three widely used methods, i.e., confidence-based methods \cite{hu2023uncertainty}, ensemble methods, and Bayesian methods. Confidence-based methods evaluate the reliability of model outputs using entropy, probability, calibration, and ensemble approaches. Ensemble methods employ multiple models to calculate the uncertainty estimation \cite{tavazza2023approaches}, while Bayesian approximation methods use computational methods \cite{lan2022scaling}, such as Monte Carlo. This section provides a brief overview of the existing literature on UQ methods applied to NLP and LLMs and provides an analysis of responses generated for a selected confusing prompt.

In the literature, there are many studies focusing on developing uncertainty quantification techniques in the context of NLP and LLMs. The study \cite{gal2016dropout} offers the use of dropout as a Bayesian approximation to estimate uncertainty in deep learning-based models. This method, known as Monte Carlo dropout, involves applying dropout during both training and inference to generate multiple stochastic forward passes and approximate the posterior distribution of the model's predictions. The research explores the use of pre-trained language models, such as BERT \cite{devlin2018bert} and GPT-3 \cite{brown2020language}, for UQ. Furthermore, most UQ studies have limited capabilities and applications, i.e., short text generation. To bridge this gap, the authors in \cite{zhang2024luq} indicate the limitations of existing methods for long text generation, and propose a novel UQ approach, called Luq-Ensemble (improved version of LUQ), which essembles responses from multiple models and selects the response with the least uncertainty. Based on the results, the proposed approach demonstrates better performance compared to traditional methods and improves the factual accuracy of the responses compared to the best individual LLM. In addition to increasing the number of applications using LLMs and high-reliability expectations, novel UQ methods and benchmarks for comprehensive evaluation strategies must be developed to understand the uncertainty of LLMs to be able to help researchers and engineers towards more robust and reliable LLMs. Fortunately, several studies propose benchmarks to address this. The authors in \cite{ye2024llm_uq} develop a benchmark for LLMs involving UQ with an uncertainty-aware evaluation metric, called UAcc for both prediction accuracy and uncertainty. The developed benchmark consists of eight LLMs (LLM series) spanning five representative natural language processing tasks. The results indicate several findings related to LLM uncertainty, i.e., higher accuracy may tend to lower certainty, and instruction fine-tuning may increase the uncertainty of LLMs.

Despite these advancements, the proposed methods may not fully capture models’ underlying uncertainties for LLMs due to several reasons mentioned earlier. Evaluating the performance of
LLMs has been essential for the development and
deployment of robust LLMs. It makes the need for novel UQ approaches more critical. One of the possible approaches is the geometric approach to UQ, particularly using convex hull analysis. This method has received limited attention in the literature. This study builds on the foundation of these existing methods and introduces a novel approach that leverages the geometric properties of response embeddings to measure uncertainty in LLM outputs. 

Table \ref{tab:example_prompt} presents an analysis of the responses generated for a strategically chosen confusing prompt: "Is a broken clock right if it tells the right time twice a day?". This prompt, carefully selected for its potential to produce a wide range of responses, is used to evaluate the variability and spread of responses at different temperature settings, i.e., 0.25, 0.5, 0.75, and 1.0, using the convex hull area as a metric.

\begin{table*}[h!]
\centering
\renewcommand{\arraystretch}{1.25}
\caption{Convex Hull Area Metrics for Responses to a Confusing Prompt}
\label{tab:example_prompt}
\begin{tabular}{|c|c|c|c|c|}
\hline 
\textbf{Prompt Value} & \multicolumn{4}{|c|}{\textbf{Is a broken clock right if it tells the right time twice a day?}} \\
\hline \hline
\multicolumn{5}{|l|}{\textbf{Response 1}} \\
\multicolumn{5}{|p{12cm}|}{If all rules have an exception, then the rule stating that all rules have an exception would also have an exception. This would mean that there is at least one rule that does not have an exception. Therefore, the statement "all rules have an exception" cannot be true.} \\
\hline
\multicolumn{5}{|l|}{\textbf{Response 2}} \\
\multicolumn{5}{|p{12cm}|}{If we consider the statement ``all rules have an exception" as a rule in itself, then the exception to this rule would be that not all rules have exceptions. This means that there are some rules that do not have exceptions, contradicting the original assertion that all rules have exceptions. Therefore, the statement "all rules have an exception" does not necessarily apply universally.} \\
\hline
\multicolumn{5}{|l|}{\textbf{Response 3}} \\
\multicolumn{5}{|p{12cm}|}{If all rules have an exception, then the rule that all rules have an exception would also have an exception. So, the exception to the rule that all rules have an exception is that not all rules have exceptions.} \\
\hline \hline
\textbf{Temperature Setting} & \multicolumn{1}{|p{1.9cm}|} {\textbf{0.25}} & \multicolumn{1}{|p{1.9cm}|} {\textbf{0.5}} & \multicolumn{1}{|p{1.9cm}|} {\textbf{0.75}} & \textbf{1.0} \\
\hline
\textbf{Convex Hull Area} & 1.3122 & 5.0236 & 6.5476 & 5.7562 \\
\hline
\end{tabular}
\end{table*}

The table includes three responses, Response 1, Response 2, and Response 3, respectively. Response 1 discusses the paradox of a rule that states all rules have exceptions, arguing that such a rule cannot be valid since it would mean that at least one rule does not have an exception. Response 2 expands on the logical inconsistency if the rule itself is an exception, suggesting that some rules must not have exceptions, which contradicts the original assertion. Response 3 is similar to Response 1, addressing the self-contradiction inherent in the statement.

In this case, the convex hull area is measured for each response (Response 1, 2, and 3) at four different temperature settings (0.25, 0.5, 0.75, and 1.0). These temperatures control the randomness in the response generation process. The convex hull area values are 1.3122 at 0.25, 5.0236 at 0.5, 6.5476 at 0.75, and 5.7562 at 1.0. These values indicate the spread and variability of the responses, which can be interpreted as follows: 
\begin{itemize}
    \item A larger convex hull area suggests greater variability, which, in the context of this study, can be indicative of increased uncertainty or diversity in the generated responses.
\end{itemize}

On the other hand, it can be considered a threshold to limit or maximize the convex hull area. As seen in the table, the areas are close to each other for high-temperature settings, i.e., 0.5, 0.75, and 1.0. It is between 5.0236 and 6.5476.

Table \ref{tab:example_prompt} also provides us with a brief analysis of how the model's responses change at different temperature settings for confusing prompts. Traditional methods can still evaluate the uncertainty of model outputs. However, they have difficulties when applied to LLMs, and novel proposed approaches can be adopted into these traditional techniques. To address this requirement, in this study, a novel geometric approach to UQ using convex hull analysis is proposed for the uncertainty of LLMs responses along with several use cases.


\section{System Model}
This section presents the proposed approach to calculate the uncertainty of the generated responses based on the geometric properties of convex hull areas for each prompt. The system processes a comprehensive set of categorized prompts, meticulously focusing on three distinct types: 'easy', 'moderate', and 'confusing'. The principal components of the model encompass the generation of embeddings utilizing BERT, dimensionality reduction via Principal Component Analysis (PCA), clustering with the Density-Based Spatial Clustering of Applications with Noise (DBSCAN) algorithm, and the computation of convex hull areas to quantify the dispersion and uncertainty within the response space.

\begin{figure*}[h]
    \centering
    \includegraphics[width=1.0\linewidth]{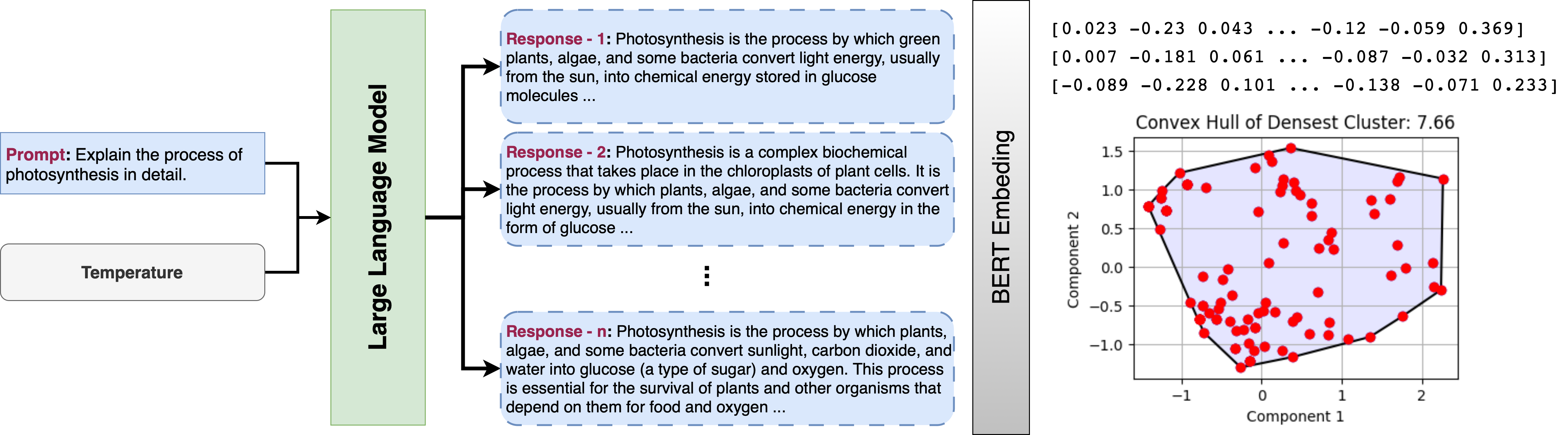}
    \caption{The system overview for calculating uncertainty in LLM responses.}
    \label{fig:system_overview}
\end{figure*}

The overview of the system is shown in Figure \ref{fig:system_overview}. The figure illustrates the workflow of the model, from the input prompt to the calculation of the convex hull area, providing a measure of uncertainty. The process begins with a given prompt (e.g., ``Explain the process of photosynthesis in detail.") and a specified temperature setting, which is fed into a LLM. The model generates multiple responses to the prompt, each of them provides a potentially different elaboration on the topic. These responses are then encoded into high-dimensional embeddings using a BERT model. The embeddings, initially in a high-dimensional space, are projected onto a two-dimensional space using Principal Component Analysis (PCA) for easier visualization and clustering. The 2D embeddings are clustered using the DBSCAN algorithm, which identifies groups of similar responses. For each cluster, a convex hull is computed, representing the smallest convex boundary that encloses all points in the cluster. The area of each convex hull is calculated, with the total area serving as a measure of the uncertainty of the model's responses to the given prompt. The example shown includes a plot of the 2D embeddings with the convex hull of the densest cluster highlighted, illustrating the spatial distribution and clustering of the responses.

Let \( \mathcal{P} = \{p_1, p_2, \ldots, p_n\} \) denote the set of prompts, which are rigorously categorized into three types: 'easy', 'moderate', and 'confusing'. For each prompt \( p \in \mathcal{P} \), it is generated a diverse set of responses \( \mathcal{R}(p) = \{r_1, r_2, \ldots, r_m\} \) using a suite of different language models \( \mathcal{M} = \{m_1, m_2, \ldots, m_k\} \) at varying temperature settings \( \mathcal{T} = \{t_1, t_2, \ldots, t_l\} \).

The embeddings for each response \( r \in \mathcal{R}(p) \) are computed utilizing a pre-trained BERT model, represented mathematically as:
\[
\mathbf{E}(r) = \text{BERT}(r)
\]
where \( \mathbf{E}(r) \in \mathbb{R}^d \) denotes the embedding vector in a \( d \)-dimensional space, encapsulating the semantic content of the response in a high-dimensional feature space.

Given the set of embeddings \( \mathbf{E}(\mathcal{R}(p)) = \{\mathbf{E}(r_1), \mathbf{E}(r_2), \ldots, \mathbf{E}(r_m)\} \), it is applied into Principal Component Analysis (PCA) to reduce the dimensionality of the embedding vectors to 2, enabling effective visualization and clustering:
\[
\mathbf{E}_\text{PCA}(\mathcal{R}(p)) = \text{PCA}(\mathbf{E}(\mathcal{R}(p)), 2)
\]
where \( \mathbf{E}_\text{PCA}(r) \in \mathbb{R}^2 \), the transformation is achieved by projecting the original high-dimensional embeddings onto a two-dimensional subspace that maximizes the variance, thereby preserving the most significant features of the data.

Subsequently, the DBSCAN algorithm is utilized to cluster the reduced embeddings to identify distinct groupings within the response space:
\[
\mathbf{L} = \text{DBSCAN}(\mathbf{E}_\text{PCA}(\mathcal{R}(p)), \epsilon = 0.25 \times t \times 4.0, \text{min\_samples}=3)
\]
where \( \mathbf{L} \) represents the set of cluster labels assigned to each embedding point, with \( \epsilon \) controlling the maximum distance between points in the same cluster and \( \text{min\_samples} \) specifying the minimum number of points required to form a cluster.

For each identified cluster \( c \in \mathbf{L} \), excluding noise points (i.e., \( c = -1 \)), we compute the convex hull and its corresponding area, which encapsulates the geometric boundary of the cluster:
\[
\text{ConvexHull}(\mathbf{E}_\text{PCA}(c)), \quad \text{Area}(\text{ConvexHull}(c))
\]
The convex hull is the smallest convex polygon that can enclose all the points in the cluster, and the area of the convex hull is a measure of the spatial extent of the cluster.

The total convex hull area for a given prompt \( p \) at temperature \( t \) is then defined as the summation of the areas of all clusters:
\[
A(p, t) = \sum_{c \in \mathbf{L}, c \neq -1} \text{Area}(\text{ConvexHull}(c))
\]
This metric provides an aggregate measure of the uncertainty and dispersion of the model's responses to the prompt, with larger areas indicating higher uncertainty.

The algorithm for calculating the uncertainty using convex hull areas is outlined in Algorithm \ref{alg:convex_hull}. This algorithm takes as input a prompt, a temperature setting, and a model, computes embeddings, reduces their dimensionality, performs clustering, and finally computes and returns the convex hull areas.

\begin{algorithm}[h]
\caption{Uncertainty Calculation Using Convex Hull Area}
\label{alg:convex_hull}
\begin{algorithmic}[1]
\REQUIRE $p$: Prompt, $m$: Model, $t$: Temperature
\STATE Initialize total hull area $\text{total\_hull\_area} \leftarrow 0$
\STATE Initialize number of clusters $\text{num\_clusters} \leftarrow 0$
\STATE $\mathcal{R}(p) \leftarrow \text{get\_responses\_for\_prompt}(p, m, t)$
\IF{$\mathcal{R}(p) = \emptyset$}
    \RETURN $0$
\ENDIF
\STATE $\mathbf{E}(\mathcal{R}(p)) \leftarrow \{\text{get\_bert\_embeddings}(r) \mid r \in \mathcal{R}(p)\}$
\STATE $\mathbf{E}_\text{PCA}(\mathcal{R}(p)) \leftarrow \text{PCA}(\mathbf{E}(\mathcal{R}(p)), 2)$
\IF{$|\mathbf{E}_\text{PCA}(\mathcal{R}(p))| < 10$}
    \RETURN $0$
\ENDIF
\STATE $\mathbf{L} \leftarrow \text{DBSCAN}(\mathbf{E}_\text{PCA}(\mathcal{R}(p)), \epsilon = 0.25 \times t \times 4.0, \text{min\_samples}=3)$
\STATE $\text{unique\_labels} \leftarrow \text{set}(\mathbf{L})$
\STATE $\text{num\_clusters} \leftarrow |\text{unique\_labels}| - (1 \text{ if } -1 \in \text{unique\_labels} \text{ else } 0)$
\FORALL{$c \in \text{unique\_labels}$}
    \IF{$c = -1$}
        \STATE \textbf{continue}
    \ENDIF
    \STATE $\text{cluster\_indices} \leftarrow \{i \mid \mathbf{L}[i] = c\}$
    \STATE $\text{cluster\_points} \leftarrow \mathbf{E}_\text{PCA}[\text{cluster\_indices}]$
    \IF{$|\text{np.unique}(\text{cluster\_points.round}(6), \text{axis}=0)| > 2$}
        \STATE $\text{hull} \leftarrow \text{ConvexHull}(\text{cluster\_points})$
        \STATE $\text{hull\_area} \leftarrow \text{hull.volume}$
        \STATE $\text{total\_hull\_area} \leftarrow \text{total\_hull\_area} + \text{hull\_area}$
    \ENDIF
\ENDFOR
\RETURN $\text{total\_hull\_area}$
\end{algorithmic}
\end{algorithm}

\section{Experimental Results and Discussion}
This section provides experimental results obtained by the convex hull-based UQ method for the selected LLMs, and a discussion on these results.
\subsection{Experimental Results}
\begin{figure}[!htbp]
    \centering
     \subfigure[ ]{
        \includegraphics[width=0.75\linewidth]{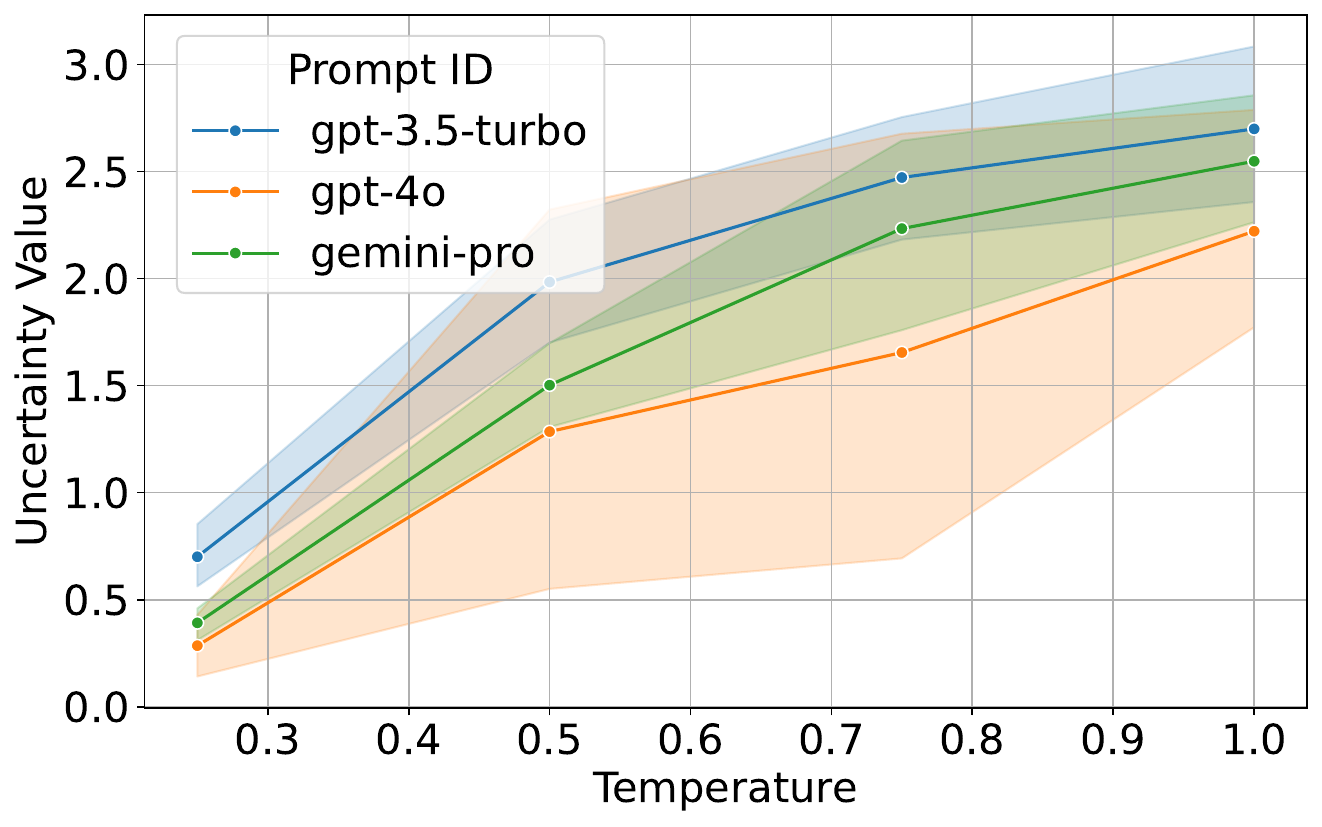}
        \label{fig:easy-unc}
    } 
    \subfigure[ ]{
        \includegraphics[width=0.75\linewidth]{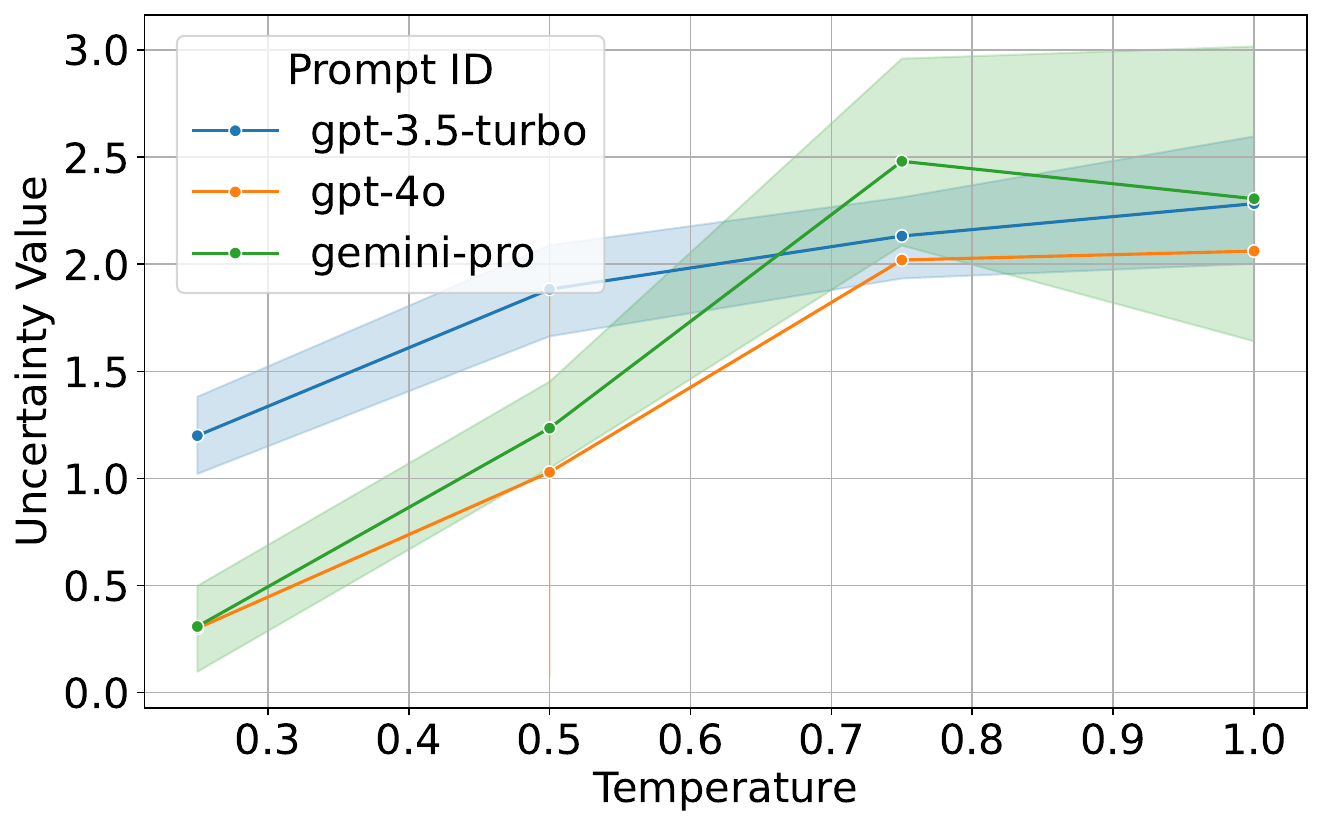}
        \label{fig:figure2}
    } \\
     \subfigure[ ]{
        \includegraphics[width=0.75\linewidth]{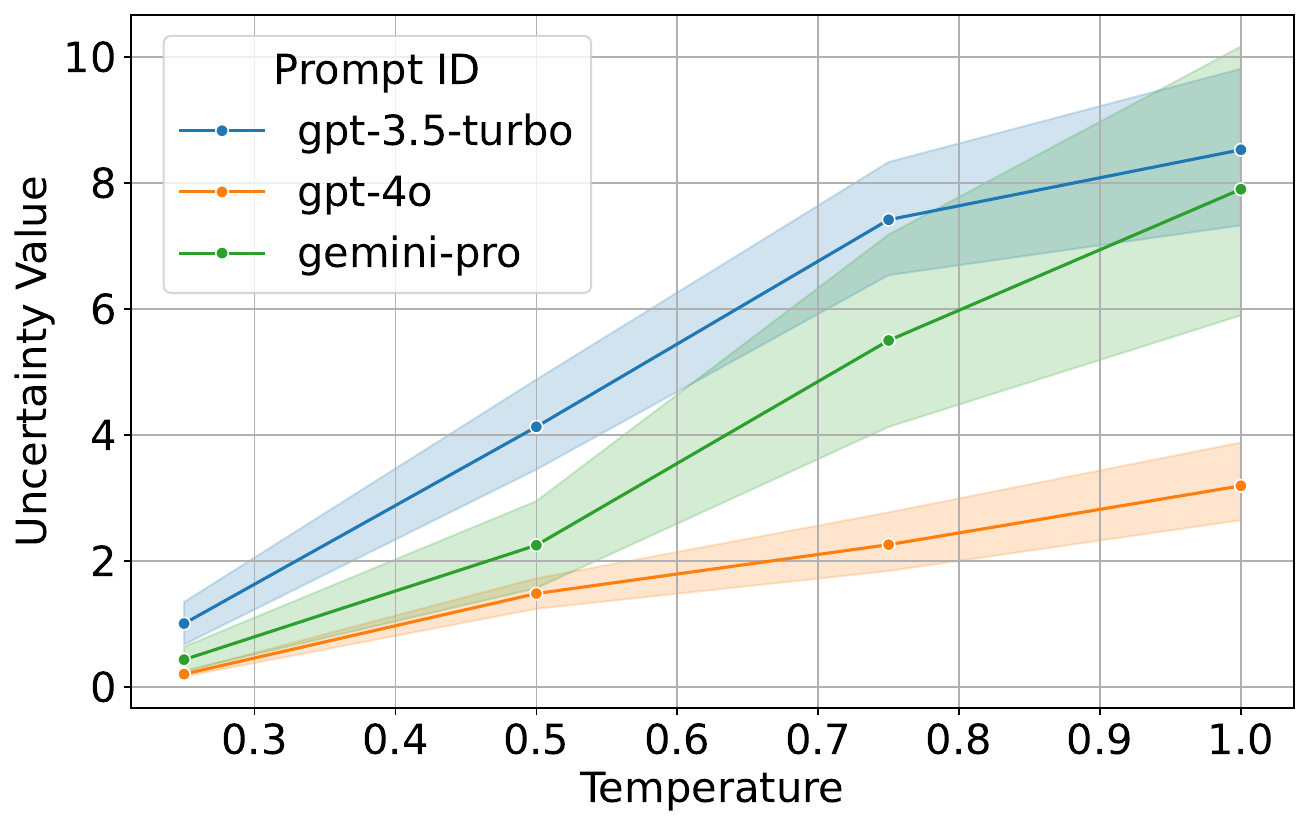}
        \label{fig:figure3}
    }
    \caption{The relationship between uncertainty and temperature settings based convex hull-based analysis for (a) easy, (b) moderate, and (c) confusing prompts of GPT-3.5-turbo, GPT-4o, and Gemini-pro outputs}
    \label{fig:unc_figures}
\end{figure}

The experiments are conducted using three types of prompts, i.e., 'easy', 'moderate', and 'confusing', with responses generated by three different models (Gemini-Pro, GPT-3.5-turbo, and GPT-4o) at various temperature settings (0.25, 0.5, 0.75, and 1.0). The results are detailed in Table \ref{tab:mean_std_unc}, \ref{tab:median_irq_unc}, and \ref{tab:clustering_analysis}, and also visualized in Figures \ref{fig:unc_figures} a-c, which plot the uncertainty values (convex hull areas) against the temperature settings for each model and prompt type.


The convex hull area measures the variability in the model responses, with higher values indicating a higher dispersion among the responses. The model and prompt type organize the table, and for each temperature setting, the mean and standard deviation of the convex hull areas are provided in each table.

Table \ref{tab:mean_std_unc} presents the statistical metrics of mean and standard deviation of the convex hull areas for different prompt types across various temperature settings for three distinct models, i.e., Gemini-Pro, GPT-3.5-turbo, and GPT-4o. According to the results for Gemini-Pro, the convex hull area for easy prompts increases from 0.3916 at a temperature of 0.25 to 2.5481 at a temperature of 1.0, and from 0.0953 at a temperature of 0.25 to 0.3782 at a temperature of 1.0 with corresponding mean and standard deviations reflecting the variability, respectively. Similar trends are observed for moderate, and confusing prompts, indicating how response variability changes with temperature settings for this model, i.e., increasing from 0.3084 at a temperature of 0.25 to 0.1.0501 at a temperature of 1.0 for moderate prompts, and from 0.4171 at a temperature of 0.25 to 7.7980 at a temperature of 1.0 for confusing prompts with corresponding mean statistical metric. GPT-3.5-turbo shows a higher mean convex hull area for confusing prompts than easy and moderate prompts, particularly noticeable at higher temperatures. For example, at a temperature of 1.0, the mean convex hull area for confusing prompts is 8.9260 with a standard deviation of 3.808616, i.e., significant response variability. Regarding GPT-4o, the mean convex hull area for easy prompts starts at 0.2851 at a temperature of 0.25 and increases to 2.2212 at a temperature of 1.0. As expected, the mean convex hull area is higher for moderate and confusing prompts. The standard deviation results increase similarly, indicating more dispersed responses at higher temperatures, i.e.,  0.0851 at a temperature of 0.25 and 1.3065 at a temperature of 1.0 for confusing prompts.



\begin{table*}[h!]
\centering 
\renewcommand{\arraystretch}{1.25}
\caption{Mean and Standard Deviation of Convex Hull Areas for Different Prompt Types and Temperatures}
\label{tab:mean_std_unc}
\begin{tabular}{|l|l|rr|rr|rr|rr|}

\hline
 & \textbf{Temperature $=>$} & \multicolumn{2}{c|}{\textbf{0.25}} & \multicolumn{2}{c|}{\textbf{0.5}} & \multicolumn{2}{c|}{\textbf{0.75}} & \multicolumn{2}{c|}{\textbf{1.0}}  \\ \cline{2-10}
 
\textbf{Model Name} & \textbf{Prompt Type} &\textbf{mean} & \textbf{std} & \textbf{mean} & \textbf{std} & \textbf{mean} & \textbf{std} & \textbf{mean} & \textbf{std}  \\ \cline{3-10}
\hline
\multirow{3}{*}{Gemini-Pro} & Easy & 0.3916 & 0.0953 & 1.5021 & 0.2357 & 2.2335 & 0.4872 & 2.5481 & 0.3782 \\
 & Moderate & 0.3084 & 0.2296 & 1.2348 & 0.2512 & 2.1244 & 0.8551 & 2.4819 & 1.0501 \\
 & Confusing & 0.4171 & 0.5205 & 2.25371 & 1.8225 & 5.4305 & 4.0952 & 7.7980 & 5.7277 \\

\cline{1-10}
\multirow{3}{*}{GPT-3.5-turbo} & Easy & 0.7003 & 0.3432 & 1.9838 & 0.7346 & 2.4720 & 0.7162 & 2.6992 & 0.9098 \\
 & Moderate & 1.1996 & 0.4416 & 1.8833 & 0.5017 & 2.1317 & 0.4254 & 2.2830 & 0.7128 \\
 & Confusing & 1.2503 & 1.0000 & 4.6584 & 1.8994 & 8.0012 & 2.8786 & 8.9260 & 3.8086 \\
\cline{1-10}
\multirow{3}{*}{GPT-4o} & Easy & 0.2851 & 0.1963 & 1.2851 & 0.9237 & 1.6546 & 0.9933 & 2.2212 & 0.5181 \\
 & Moderate & 0.3003 & 0.0012 & 1.0289 & 1.3489 & 2.0197 & 0.0012 & 2.0614 & 0.0003 \\
 & Confusing & 0.2015 & 0.0851 & 1.4806 & 0.5108 & 2.2565 & 0.9760 & 3.1909 & 1.3065 \\

\cline{1-10}
\hline
\end{tabular}
\end{table*}

Table \ref{tab:median_irq_unc} provides a comprehensive analysis of the median convex hull areas and the Interquartile Range (IQR) for different prompt types at various temperature settings for selected language models, i.e., Gemini-Pro, GPT-3.5-turbo, and GPT-4o, as in the previous case. The median statistical metric in this table represents the central tendency of the convex hull areas, indicating the typical response variability for each combination of model, prompt type, and temperature. The IQR, calculated as the difference between the 75th percentile and the 25th percentile, quantifies the spread of the middle 50\% of the convex hull areas, which provides understanding of the consistency and variability of the model responses. In this case, the results are also analyzed for each selected LLM. For Gemini-Pro, the median convex hull areas increase with higher temperatures across all prompt types. For example, the median value for confusing prompts rises from 0.1625 at 0.25 to 6.9240 at a temperature of 1.0. The corresponding IQR values indicate substantial variability, particularly at higher temperatures, i.e., the model's responses become more diverse as temperature increases, i.e., IQR value for confusing prompts rises from 0.5654 at 0.25 to 6.9493 at a temperature of 1.0. The GPT-3.5-turbo model shows exceptionally high median convex hull areas for confusing prompts, especially at higher temperatures. At a temperature of 1.0, the median convex hull area for confusing prompts reaches 9.4878, with an IQR of 4.9531, highlighting significant variability and response spread. The GPT-3.5-turbo model also shows significant increases in variability across other prompt types at high temperatures. For GPT-4o, the median convex hull areas show a reasonable increase with temperature in all prompt types. For easy prompts, the median value increases from 0.2933 at a temperature of 0.25 to 2.1034 at a temperature of 1.0. The IQR values for confusing prompts are notable, indicating increased response variability at higher temperatures as expected, i.e., increasing from 0.1412 at a temperature of 0.25 to 1.2053 at a temperature of 1.0.

Tables \ref{tab:mean_std_unc} and \ref{tab:median_irq_unc} provide comprehensive results understanding how temperature settings and prompt types influence the variability and consistency of each model response. By examining the median and IQR values, it can be gained for robust and reliable mode responses under different conditions in terms of the prompt type and temperature.

\begin{table*}[h!]
\centering 
\renewcommand{\arraystretch}{1.25}
\caption{Median and Interquartile Range (IQR) of Convex Hull Areas for Different Prompt Types and Temperatures}
\label{tab:median_irq_unc}
\begin{tabular}{|l|l|rr|rr|rr|rr|}

\hline
 & \textbf{Temperature $=>$} & \multicolumn{2}{c|}{\textbf{0.25}} & \multicolumn{2}{c|}{\textbf{0.5}} & \multicolumn{2}{c|}{\textbf{0.75}} & \multicolumn{2}{c|}{\textbf{1.0}}  \\ \cline{2-10}
 
\textbf{Model Name} & \textbf{Prompt Type} &\textbf{Median} & \textbf{IQR} & \textbf{Median} & \textbf{IQR} & \textbf{Median} & \textbf{IQR} & \textbf{Median} & \textbf{IQR}  \\ \cline{3-10}
\hline
\multirow[t]{3}{*}{Gemini-pro} & Easy & 0.4532 & 0.1270 & 1.5043 & 0.3349 & 2.2566 & 0.4168 & 2.5394 & 0.4085 \\
 & Moderate & 0.3303 & 0.1859 & 1.1918 & 0.2734 & 2.1185 & 0.9370 & 2.2959 & 0.9755 \\
 & Confusing & 0.1625 & 0.5654 & 1.7021 & 3.1887 & 4.8805 & 5.4769 & 6.9240 & 6.9493 \\

\cline{1-10}
\multirow[t]{3}{*}{GPT-3.5-turbo} & Easy & 0.6593 & 0.4452 & 1.8303 & 1.1436 & 2.3268 & 0.9757 & 2.5421 & 0.7476 \\
 & Moderate & 1.1167 & 0.7331 & 1.8942 & 0.3902 & 2.1663 & 0.5371 & 2.2770 & 0.5775 \\
 & Confusing & 1.0323 & 1.5207 & 4.2559 & 2.4682 & 8.1903 & 4.4048 & 9.4878 & 4.9531 \\
\cline{1-10}
\multirow[t]{3}{*}{GPT-4o} & Easy & 0.2933 & 0.3015 & 0.9827 & 0.8858 & 1.5944 & 0.9919 & 2.1034 & 0.5079 \\
 & Moderate & 0.3003 & 0.0000 & 1.0289 & 0.9538 & 2.0197 & 0.0000 & 2.0614 & 0.0000 \\
 & Confusing & 0.2121 & 0.1412 & 1.5880 & 0.6366 & 2.0519 & 0.8056 & 2.7609 & 1.2053 \\

\cline{1-10}
\hline
\end{tabular}
\end{table*}

Table \ref{tab:clustering_analysis} presents a comprehensive clustering analysis for the responses generated from three models, i.e., Gemini-Pro, GPT-3.5-turbo and GPT-4o, and three different prompt types in terms of mean and standard deviation statistical metrics. The key findings of this analysis are the average number of clusters (mean and standard deviation), the average area of these clusters (mean and standard deviation), and the standard deviation of the cluster areas themselves (mean and standard deviation). For the Gemini-Pro model, the number of clusters varies with the prompt type. Confusing prompts result in a mean of 2.2413, with a standard deviation of 1.3270 for the average number of clusters, indicating high variability in clustering. The average cluster area for confusing prompts is 6.6474, with a standard deviation of around 5.6263, i.e., substantial diversity in response clustering. Easy prompts lead to fewer clusters (1.5142), with a more consistent cluster area, as shown by the lower standard deviations (0.8868). The GPT-3.5-turbo model demonstrates a highly consistent clustering response for all prompt types, forming exactly one cluster for each and moderate prompt types (mean of 1.00 and standard deviation of 0.00). Despite this, the average cluster area for confusing prompts is the highest at 8.8320, with a significant standard deviation of 3.9291, indicating that while only one cluster is formed, the size of this cluster can vary greatly. For the GPT-4o model, the clustering results are relatively stable, averaging slightly over one cluster for easy and moderate prompts. The average cluster area for confusing prompts is 2.9049, with a standard deviation of 1.3908, indicating moderate variability in response clustering. The standard deviation of the cluster areas remains low in all prompt types, offering a more consistent cluster size.

Table \ref{tab:clustering_analysis} highlights how different models handle response clustering for various prompts, and illustrates how the number and size of clusters vary with prompt complexity and model type.

\begin{table*}[h!]
\centering
\renewcommand{\arraystretch}{1.25}
\caption{Clustering Analysis of Model Responses for Different Prompt Types}
\label{tab:clustering_analysis}
\begin{tabular}{|l|l|rr|rr|rr|}
\hline
 &  & \multicolumn{2}{r|}{\textbf{Num Clusters}} & \multicolumn{2}{r|}{\textbf{Cluster Area Mean}} & \multicolumn{2}{r|}{\textbf{Cluster Area Std}} \\ \cline{3-8}

\textbf{Model Name} & \textbf{Prompt Type}  & \textbf{mean} & \textbf{std} & \textbf{mean} & \textbf{std} & \textbf{mean} & \textbf{std}   \\ \cline{3-8}

\hline
\multirow[t]{3}{*}{Gemini-pro} & Easy & 1.5142 & 0.8868 & 1.7960 & 1.1820 & 0.0233 & 0.0403 \\
 & Moderate & 1.0000 & 0.0000 & 2.4652 & 1.1440 & 0.0000 & 0.0000 \\
 & Confusing & 2.2413 & 1.3270& 6.6474 & 5.6263 & 0.9985 & 2.1574 \\
\cline{1-8}
\multirow[t]{3}{*}{GPT-3.5-turbo} & Easy & 1.0000 & 0.0000 & 2.7027 & 0.7564 & 0.0000 & 0.0000 \\
 & Moderate & 1.0000 & 0.0000 & 2.3186 & 0.7912 & 0.0000 & 0.0000 \\
 & Confusing & 1.0689 & 0.2578 & 8.8320 & 3.9291 & 0.0450 & 0.2424 \\
\cline{1-8}
\multirow[t]{3}{*}{GPT-4o} & Easy & 1.2285 & 0.6456 & 1.9020 & 1.1219 & 0.0099 & 0.0279 \\
 & Moderate & 1.0740 & 0.2668 & 2.7973 & 0.9873 & 0.0015 & 0.0057 \\
 & Confusing & 1.0689 & 0.2578 & 2.9049 & 1.3908 & 0.0769 & 0.2876 \\
\cline{1-8}
\hline
\end{tabular}
\end{table*}

\begin{figure*}[htbp]
    \centering
    \subfigure[ ]{
        \includegraphics[width=0.45\linewidth]{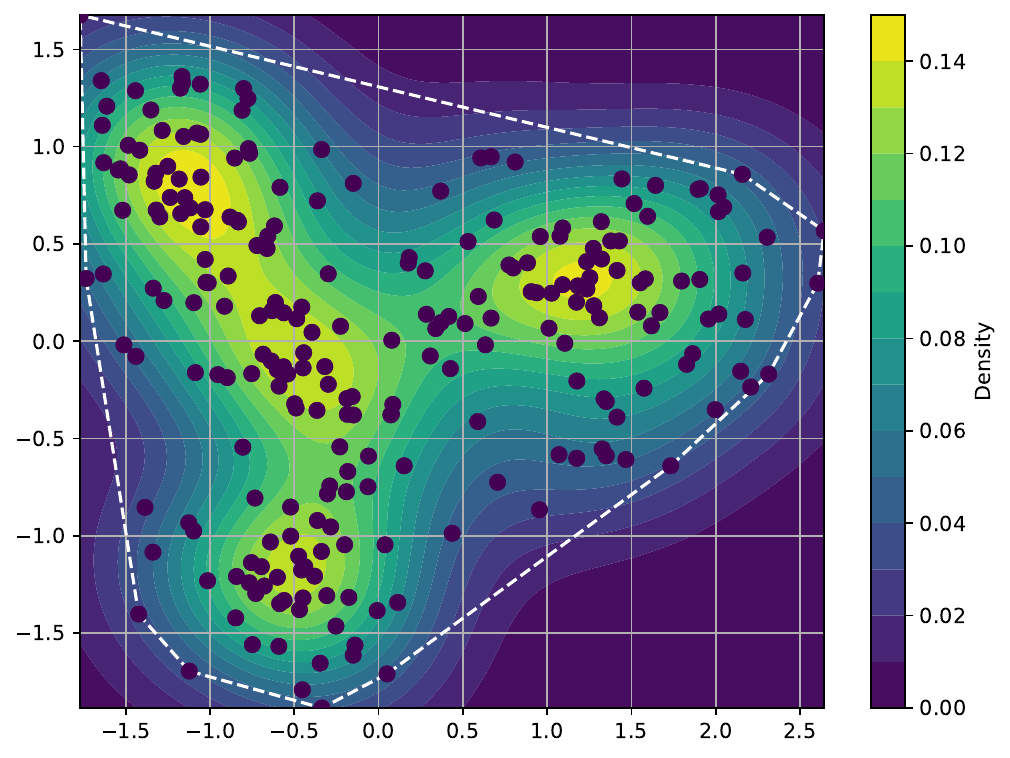}
        \label{fig:subfig1}
    }
    \hspace{0.05\textwidth}
    \subfigure[]{
        \includegraphics[width=0.45\linewidth]{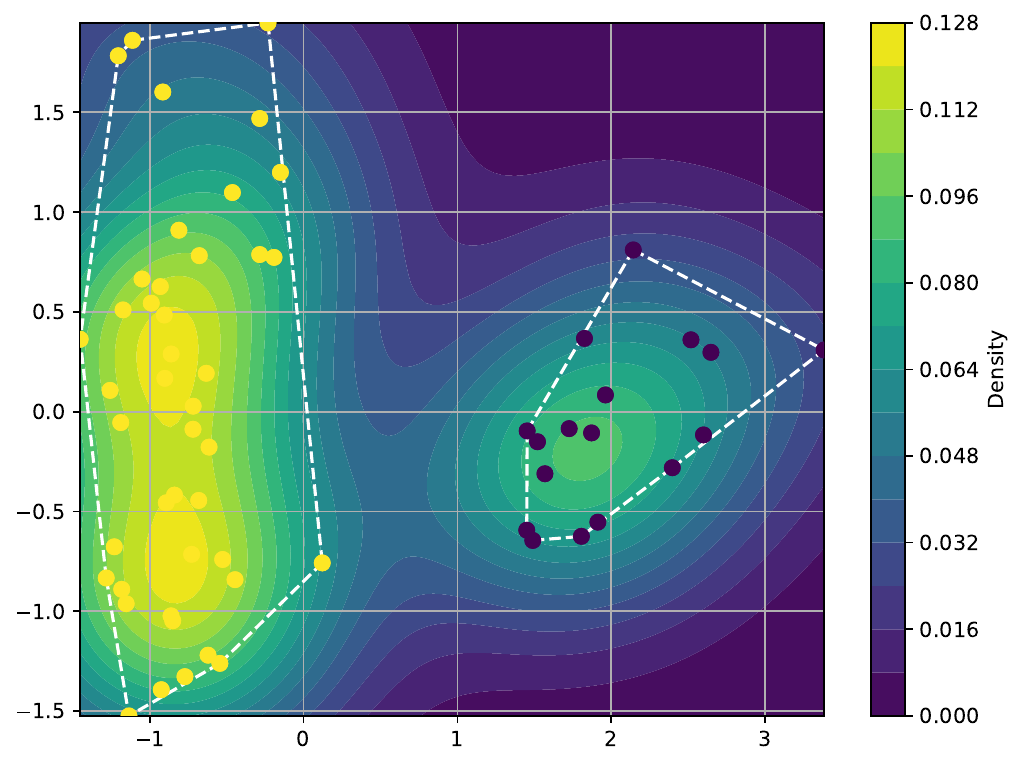}
        \label{fig:subfig2}
    }
    \caption{(a) Convex hull analysis for a prompt with a single cluster, indicating low variability and greater certainty in the model's responses. (b) Convex hull analysis for a prompt with several clusters, indicating high variability and uncertainty in the model's responses.}
    \label{fig:example_clusters}
\end{figure*}

Convex hull analysis is performed on the responses generated by the model for two different prompts at a temperature setting of 1.0. The visualizations of the convex hulls for these prompts are shown in Figure \ref{fig:example_clusters}. Each subfigure illustrates the two-dimensional projection of the response embeddings after applying Principal Component Analysis (PCA) and the clustering of these embeddings using the DBSCAN algorithm.

Figure \ref{fig:subfig1} presents the convex hull analysis for a prompt that resulted in a single dominant cluster. In this case, the model responses were more homogeneous, converging into one main group with little variability. The single cluster with a smaller convex hull area suggests that the prompt was straightforward or unambiguous, leading the model to generate more consistent responses. This lower variability indicates greater certainty in the model's outputs for this particular prompt.

Subfigure \ref{fig:subfig2} shows the convex hull analysis for a different prompt that resulted in several distinct clusters. This prompt led the model to produce a diverse set of responses that could be grouped into multiple clusters, each represented by a separate convex hull. The presence of multiple clusters indicates significant variability in the model's responses, suggesting that the prompt was either ambiguous or complex, allowing for multiple valid interpretations. The larger overall convex hull area reflects the higher uncertainty in the model's outputs, as the responses are spread across a wider semantic space.

These two examples illustrate the effectiveness of our convex hull-based approach in capturing the nuances of model uncertainty. By analyzing the number of clusters and the areas of their convex hulls, we can quantify the dispersion and variability in model responses, providing a robust measure of uncertainty.

\subsection{Discussion}
This subsection investigates the results for each prompt type, i.e., 'Easy', 'Moderate', and 'Confusing', in Figures \ref{fig:unc_figures}a-c, in terms of uncertainty value and temperature setting.
\subsubsection{Easy Prompts}

Figure \ref{fig:unc_figures}(a) illustrates the relationship between the uncertainty values and the temperature settings for easy prompts. As observed, the uncertainty values tend to increase with higher temperatures in all models. It is an expected pattern for LLMs and their responses. This trend indicates that as the temperature increases, the responses become more diverse, leading to larger convex hull areas. Among the three models, GPT-3.5-turbo exhibits the highest uncertainty at higher temperatures, i.e., a mean of 2.2692 at 1.0, followed by Gemini-pro (a mean of 2.5481 at 1.0) and GPT-4o (mean of 2.2212 at 1.0). This indicates that GPT-4o shows the lowest uncertainty at lower temperature settings, i.e., 0.2851 at 0.25, and generates a broader range of responses under varying temperature settings, reflecting its ability to capture a wider spectrum of possible outputs.


\subsubsection{Moderate Prompts}

The results for moderate prompts, shown in Figure \ref{fig:unc_figures}(b), follow a pattern similar to those of easy prompts, with increasing uncertainty values as the temperature increases. However, the overall uncertainty values are higher for moderate prompts compared to easy prompts, reflecting the increased complexity and variability of LLMs' responses. Gemini-Pro shows the highest uncertainty, particularly at high temperatures, i.e., a mean of 2.4819 at 1.0. GPT-3.5-turbo and GPT-4o also demonstrate increasing uncertainty with higher temperatures, a mean of 2.2830 at 0.25 and a mean of 2.0614 at 1.0. All models tend to low uncertainty values at low-temperature settings as expected, i.e., a mean of 0.3084, 1.1996, and 0.3003 at 0.25 for Gemini-Pro, GPT-3.5-turbo, and GPT-4o, respectively. However, GPT-3.5-turbo shows a higher uncertainty value (a mean of 1.1996) at a temperature setting of 0.25 compared to the other LLMs.


\subsubsection{Confusing Prompts}

Figure \ref{fig:unc_figures}(c) presents the uncertainty values for confusing prompts. As expected, the uncertainty is significantly higher for confusing prompts compared to easy and moderate prompts, highlighting the challenge of generating consistent responses to complex and ambiguous queries. The increase in uncertainty with temperature is expected for confusing prompts, particularly for GPT-3.5-turbo and Gemini-pro at higher temperatures, i.e., a mean of 8.9260, and 7.7980 at 1.0. GPT-4o offers better performance, i.e, a mean of 3.1909 at 1.0, and tends to a more gradual increase (between 0.2015 and 3.1909 at 0.5 and 1.0 temperature settings, respectively) compared to the other two models. This indicates that GPT-3.5-turbo and Gemini-pro are more sensitive to high-temperature settings, generating a wider range of responses when faced with confusing prompts. 


\section{Observations}


The results from these experiments demonstrate the importance of the proposed convex hull-based approach in UQ of LLMs' responses. The convex hull areas provide a robust measure of the diversity and variability of LLMs' responses, with a higher value indicating greater uncertainty. The analysis for the selected cases can be extended through the detailed results. However, the following observations can be highlighted and categorized into temperature setting, LLM comparison, and prompt complexity:

\begin{enumerate}
    \item \textbf{Temperature Setting:} All models show increasing uncertainty with higher temperature settings, i.e., over 0.75. It is a highly expected observation. It highlights the importance of temperature settings in controlling the randomness and diversity of generated LLMs' responses.
    \item \textbf{LLM Comparison:} GPT-4o consistently offers a better performance in terms of uncertainty compared to GPT-3.5-turbo and Gemini-pro. On the other hand, GPT-3.5-turbo shows worse performance, particularly at higher temperature settings, i.e., generating more diverse responses.
   \item \textbf{Prompt Complexity:} The uncertainty values are higher for more complex prompts, i.e., confusing prompts, across all selected models, indicating the inherent difficulty in generating consistent responses to such queries.
\end{enumerate}

These observations emphasize the importance of temperature settings and prompt complexity in the uncertainty of LLMs' responses. The proposed convex-hull-based geometric approach provides a novel and effective metric to capture the uncertainty of LLMs' responses, as well as valuable perception for the development and evaluation of current and future LLMs.

\section{Conclusion}
This study proposes a novel geometric approach to quantifying uncertainty for LLMs. The proposed approach utilizes the spatial properties of convex hulls formed by embeddings of model responses to measure dispersion, particularly complex and high-dimensional text generated by LLMs. In this study, three types of prompts are used, i.e., 'easy', 'moderate', and 'confusing', for three different LLMs, namely GPT-3.5-turbo, GPT-4, and Gemini-pro at various temperature settings, i.e., 0.25, 0.5, 0.75, and 1.0. The responses are transformed into high-dimensional embeddings using a BERT model and then projected into a 2D space using PCA. DBSCAN algorithm is utilized to identify clusters. The convex hull areas that surround these clusters were used as our metric of uncertainty. The proposed approach provides a clear and interpretable metric to develop more reliable LLMs. By understanding the factors that influence uncertainty, such as temperature settings and prompt complexity, researchers and engineers can develop and evaluate their LLMs in a better way. The experimental results indicated several key findings: (1) the uncertainty increased with higher temperature settings across all models, highlighting the role of temperature in controlling the diversity of generated responses, (2) GPT-4 consistently exhibited higher uncertainty values compared to GPT-3.5-turbo and Gemini-pro, indicating its enhanced ability to produce a wider range of responses, and (3) the uncertainty values were significantly higher for more complex prompts, particularly confusing prompts, across all models, reflecting the inherent challenge in generating consistent responses to such queries.



For future work, the proposed geometric approach will be extended to other types of embedding and clustering algorithms for more robust and reliable LLMs in critical domains, such as healthcare, finance, and law. In addition, integrating this uncertainty metric with other evaluation criteria, such as accuracy and coherence, could provide a more comprehensive assessment of LLM performance. 

\bibliographystyle{IEEEtran}
\bibliography{ref}

\end{document}